\documentclass[letterpaper, 10 pt, conference]{ieeeconf}
\IEEEoverridecommandlockouts
\usepackage[utf8]{inputenc}
\usepackage[T1]{fontenc}
\usepackage{graphicx}
\usepackage{comment}
\usepackage{amsmath}
\usepackage{cleveref}
\usepackage{array}
\usepackage{xcolor}
\usepackage{multirow}

\graphicspath{ {./figures/} }

\title{\LARGE \bf
Results and Lessons Learned from Autonomous Driving Transportation Services in Airfield, Crowded Indoor, and Urban Environments
}

\author{Doosan Baek$^{1}$,
Sanghyun Kim$^{1,2}$,
Seung-Woo Seo$^{1}$,
and
Sang-Hyun Lee$^{1}$
\thanks{This work is supported by the Institute of New Media and Communications, and the Automation and Systems at Seoul National University and ThorDrive, Co., Ltd. (\textit{Corresponding author: Sang-Hyun Lee})}
\thanks{$^{1}$Seoul National University, Korea
        (e-mail: \{doosan528, shyun613, sseo, slee01\}@snu.ac.kr)}
\thanks{$^{2}$ThorDrive, Korea
        (e-mail: \{shkim\}@thordrive.ai)}
}

\begin{document}
\maketitle
\thispagestyle{empty}
\pagestyle{empty}

\begin{abstract}
Autonomous vehicles have been actively investigated over the past few decades. Several recent works show the potential of autonomous vehicles in urban environments with impressive experimental results. However, these works note that autonomous vehicles are still occasionally inferior to expert drivers in complex scenarios. Furthermore, they do not focus on the possibilities of autonomous driving transportation services in other areas beyond urban environments. This paper presents the research results and lessons learned from autonomous driving transportation services in airfield, crowded indoor, and urban environments\footnotemark[3]{}. We discuss how we address several unique challenges in these diverse environments. We also offer an overview of remaining challenges that have not received much attention but must be addressed. This paper aims to share our unique experience to support researchers who are interested in exploring autonomous driving transportation services in various real-world environments.
\end{abstract}


\footnotetext[3]{Videos for our autonomous driving services are available at https://www.youtube.com/@thordrive3021}

\section{Introduction}
Recently, academia and the industry have paid much attention to the great potential of autonomous vehicles. Furthermore, the public’s interest and understanding of autonomous vehicles have rapidly increased in the past few years as product-level autonomous vehicles emerge beyond the research stage and the public can now directly experience autonomous driving. Many commercial vehicles have a level-2 autonomous driving system defined by the Society of Automotive Engineers \cite{borrego2020resource}. Particularly, Tesla’s full self-driving (FSD) allows people to experience the autonomous driving system in daily life through their vehicles.

\begin{figure}[tp]
    \centering
    \includegraphics[width=0.95\columnwidth]{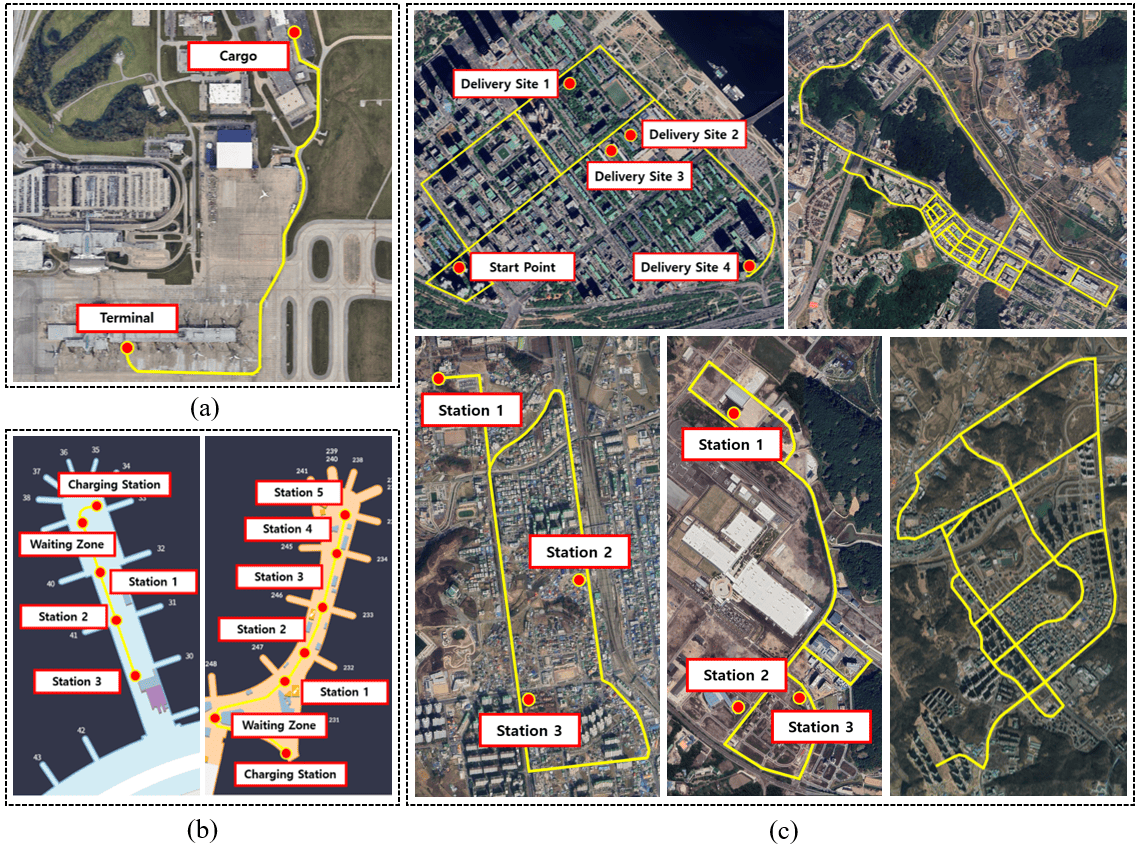}  
    \vskip -0.1in
    \caption{    
    Autonomous driving service area. (a) Airfields: Cincinnati/Northern Kentucky Airport (2.68 km). (b) Crowded indoors: Incheon Airport Terminal 1 (300 m) and Terminal 2 (550 m). (c) Urban: top: from the left, last-mile delivery in Seoul (4.67 km) and testbed in Seoul (8.21 km) and bottom: from the left, shuttle in Sejong (6.12 km), health care service in Gwangju (5.2 km), and road infra monitoring in Hwaseong (15.5 km).
    }
    \label{fig:environment}
    \vskip -0.2in
\end{figure}

Contrary to public expectations, several recent works have shown that current autonomous vehicles are occasionally inferior to expert drivers in complex urban driving scenarios. These vehicles sometimes generate awkward driving behaviors in unsignalized intersection or roundabout scenarios as they struggle to interpret interactions with surrounding vehicles. Such awkward driving behaviors can cause traffic congestion or obstruct the routes of emergency vehicles. These circumstances indicate that there are still many challenges remaining to fully realize the benefits of autonomous driving transportation services in urban environments \cite{burnett2021zeus, milanes2021tornado}.

Autonomous vehicles can be applied to diverse transportation services beyond urban driving.
Tow tractors that transport baggage and cargo at airfields are receiving much attention because of labor shortages and rising labor costs \cite{morris2015self}. Compact vehicles for passengers with reduced mobility have been developed for indoor environments, such as large shopping malls \cite{morales2013human, leaman2017comprehensive}. Applying autonomous vehicles to these services can have a significant positive effect on our society. However, it presents unique and challenging problems that are not encountered in urban environments. Furthermore, these problems have not yet been sufficiently considered in previous works.

\begin{figure*}[t]
    \centering
    \includegraphics[width=0.89\textwidth]{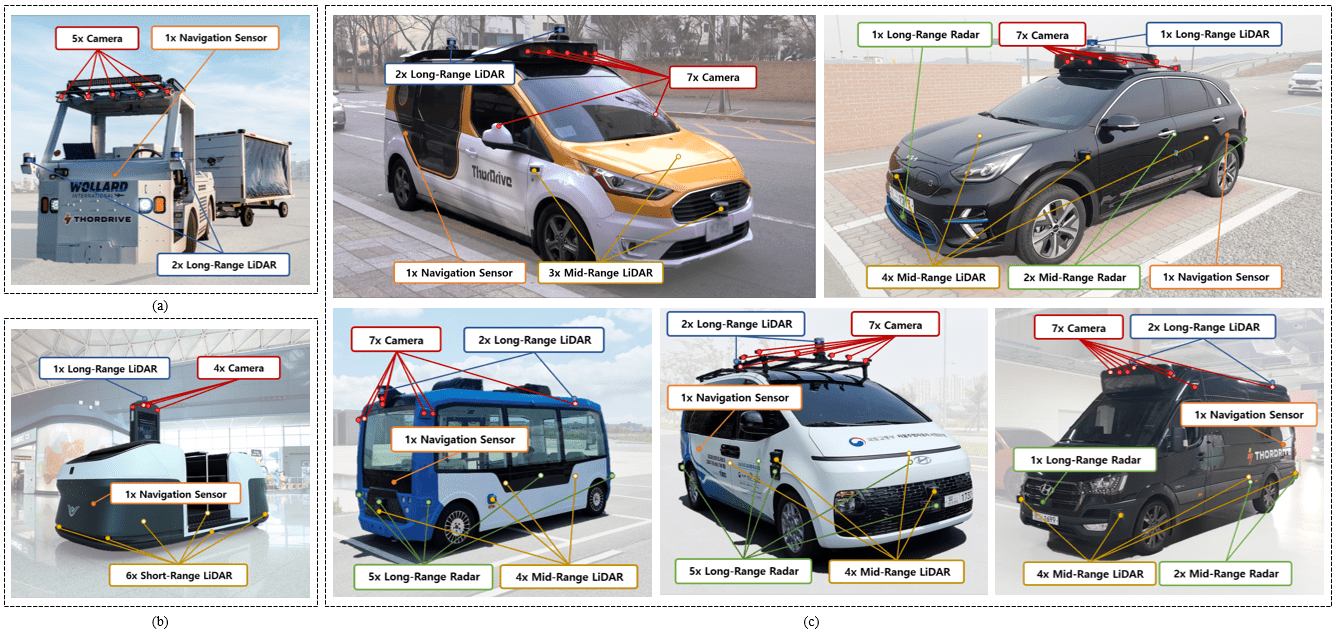}
    \vskip -0.1in
    \caption{
    Developed autonomous driving platforms. (a) Airfields: tow tractor. (b) Crowded indoors: shuttle (AirRide). (c) Urban: top: from the left, last-mile delivery vehicle (Eligo) and on-demand taxi, and bottom: from the left, health care service vehicle, road infra monitoring vehicle, and shuttle.
    }
    \label{fig:vehicle platform}
    \vskip -0.15in
\end{figure*}

In this paper, we present the results and lessons learned from autonomous driving transportation services in airfield, crowded indoor, and urban environments. Figs. \ref{fig:environment} and \ref{fig:vehicle platform} describe our service areas and vehicle platforms. In particular, we ran autonomous tow tractor services at Cincinnati/Northern Kentucky International Airport (CVG) airfield. We have also been operating indoor autonomous shuttle services at Incheon International Airport (ICN) arrival and departure halls. Lastly, we have been operating last-mile delivery, shuttle, and taxi services in urban environments, including Seoul, Sejong, and Jinhae in South Korea.

To successfully complete our autonomous driving services, we propose novel and general algorithms to handle their unique challenges. We also discuss other remaining challenges and explore promising research directions for addressing them. These challenges, including jetblast detection and inevitable traffic rule violation, have not received much attention but must be addressed to realize product-level autonomous vehicles. Our work is in contrast to most previous works that focused only on autonomous driving transportation services in urban environments or dealt less with technical issues from a service perspective.

We aim to share our unique experience to support researchers who are interested in deploying autonomous vehicles in diverse environments. The main contributions of our work are summarized as follows:
1) we successfully conducted autonomous driving transportation services in airfield, crowded indoor, and urban environments, 2) we present how we address diverse and unique challenges for each service, and 3) we discuss the remaining challenges that have not yet received much attention but must be resolved for product-level autonomous vehicles.

\section{Related Works}

\begin{figure*}[ht]
    \centering
    \includegraphics[width=0.85\textwidth]{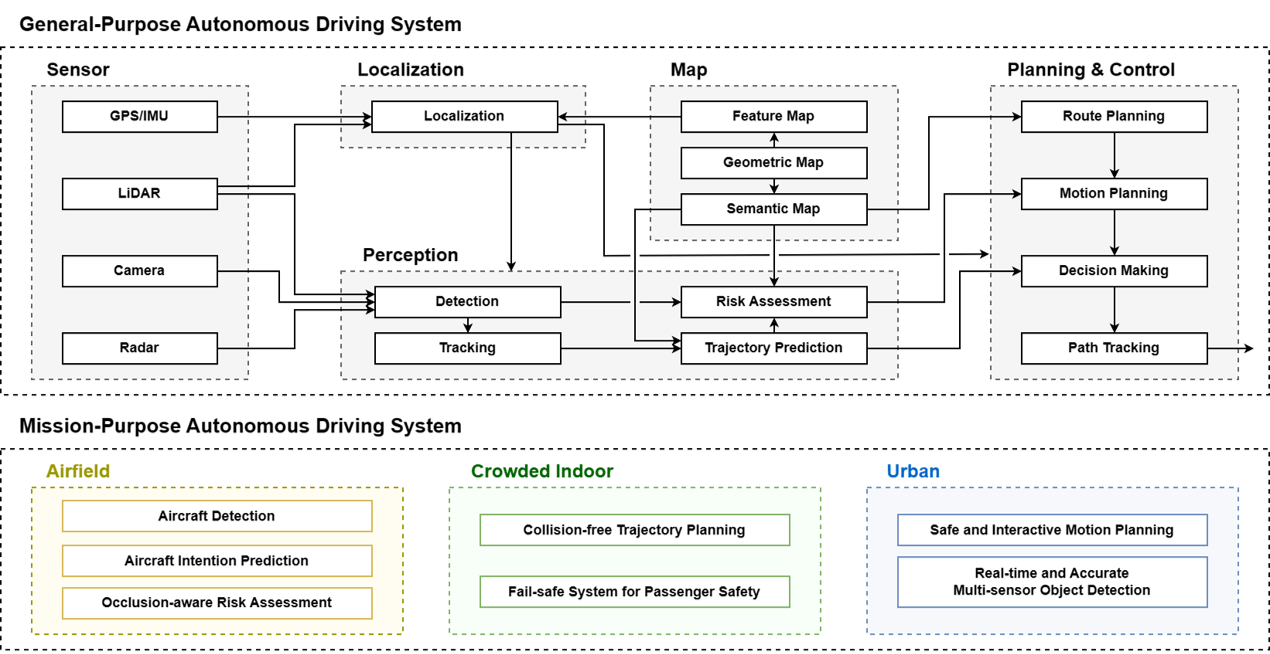}
    \vskip -0.1in
    \caption{Overall architecture of the general- and mission-purpose autonomous driving system.}
    \label{fig:system architecture}
    \vskip -0.2in
\end{figure*}

Many impressive demonstrations have shown that autonomous vehicles can handle diverse driving scenarios in urban environments \cite{nothdurft2011stadtpilot, broggi2015proud, ziegler2014making}. Nothdurft \textit{et al.} \cite{nothdurft2011stadtpilot} introduced one of the first autonomous driving tests in urban environments. Their autonomous vehicle handled lane keeping, signalized intersections, and lane change scenarios in Braunschweig. Broggi \textit{et al.} \cite{broggi2015proud} demonstrated autonomous driving on public urban roads in Parma and addressed challenging urban driving scenarios, such as roundabouts, junctions, and pedestrian crossings. In the paper of Ziegler \textit{et al.} \cite{ziegler2014making}, an autonomous vehicle equipped with close-to-market sensors completed 103 km of the Bertha Benz Memorial Route, including diverse urban areas. Unlike these works that conducted autonomous driving tests only in urban environments, our work conducted autonomous driving services in more realistic scenarios across diverse environments, including airfields and crowded indoors.

As a solution to labor shortages and rising labor costs, there is a growing focus on autonomous ground support equipment (GSE) in airfields. Wang \textit{et al.} \cite{wang2021accurate} introduced an efficient dispatching framework for a fleet management system of GSE. Morris \textit{et al.} \cite{morris2015self} presented an autonomous push-back tractor service, in which the push-back tractor tows aircraft from the gate to the runway and from the runway to the gate. The experimental results of these works demonstrate that autonomous GSE can improve the efficiency of operations in airfields. However, these works do not present in-depth discussions about the challenges of autonomous driving in the airfields. Unlike these works, we propose novel perception and motion planning algorithms to handle the challenges of airfield autonomous driving and provide several future research directions.

Another major application of autonomous driving is indoor robots. Lee \textit{et al.} \cite{lee2023fusionloc} introduced relocalization algorithms for serving robots that often fail to estimate their pose. Tsiogas \textit{et al.} \cite{tsiogas2021pallet} demonstrated the autonomous pallet-moving vehicle in the factory, focusing on pallet detection and docking scenarios. Kastner \textit{et al.} \cite{kastner2022human} presented human-following and -guiding service robots. They proposed a semantic deep reinforcement learning algorithm that enables robots to navigate safely in crowded environments. While these works have shown various indoor services with robots, our work discusses another service: indoor autonomous shuttles in airports. The vehicle used for this service is larger than typical indoor robots to accommodate multi-passengers and ensures their safety and comfort, which leads to distinct challenges and research directions.

\section{Autonomous Driving Transportation Services in Various Challenging Environments}
Our autonomous driving system is a combination of a general-purpose system that is agnostic to environments and a mission-purpose system that is designed to address environment-specific problems (Fig. \ref{fig:system architecture}). The general-purpose system is largely divided into localization, perception, planning, and control modules. The localization module estimates the current location of the ego vehicle and the perception module detects surrounding objects and predicts their behaviors. The planning module generates the optimal trajectory based on the results of both localization and perception modules. The control module estimates the control values to follow the optimal trajectory. The mission-purpose system comprises the novel algorithms proposed in this work. In the remainder of this section, we discuss how the proposed algorithms in the mission-purpose system handle diverse challenges for each environment in detail.

\subsection{Airfield Environment}
We operated an autonomous tow tractor for transporting passenger baggage and cargo within a 2.68 km-long airfield at the CVG. The airfield environment has unique objects, such as aircraft, tow tractors, and K-loaders, and each object has a different driving strategy. Figs. \ref{fig:environment}a and \ref{fig:vehicle platform}a show our tow tractors and the corresponding routes, respectively.
We conducted a performance evaluation over three weeks, with a total distance of 2,163.2 km and an operation time of 106.7 h. Table \ref{table:airfield test statistics} presents the results of the performance evaluation in the third week, where "diseng." denotes disengagement.

\begin{table}[hbt!]
    \centering
    \caption{Autonomous Tow Tractor Performance Evaluation.}
    \label{table:airfield test statistics}
    \vskip -0.05in
    \renewcommand{\arraystretch}{1.1}

    \resizebox{\columnwidth}{!}{%
    \footnotesize
        \begin{tabular}{c | c | c | c | c}
            \hline 
            \multicolumn{3}{c|}{Number of disengagements} &
            \multirow{2}{*}{
                \begin{tabular}[c]{@{}l@{}}
                    Zero diseng. \\ 
                    mission rates
                \end{tabular}
            } &
            \multirow{2}{*}{
                \begin{tabular}[c]{@{}l@{}}
                    km per \\ 
                    diseng.
                \end{tabular}
            } \\ 
            \cline{1-3}
            
            \multicolumn{1}{c|}{Safety} & 
            \multicolumn{1}{c|}{Emergency} & 
            Non-safety & & \\
            \hline
            
            11 & 1 & 51 & 81.3 \% & 13.248 \\
            \hline
        \end{tabular}%
    }
    \vskip -0.05in

\end{table}

\vskip 1mm
\subsubsection{Sensor Fusion for Aircraft Detection}
With its large sizes and irregular shapes, aircraft are one of the most difficult objects to detect in airfields. Most previous aircraft detection algorithms assume access to top-view images \cite{chen2018end}. However, autonomous GSE typically cannot utilize such images in real-time, and aircraft detection by using the sensors equipped with autonomous GSE has not been addressed yet. We also empirically observed that detection algorithms for urban driving scenarios exhibited low accuracy when applied to aircraft due to their distinct size and shape characteristics.

To overcome the limitations of previous algorithms, we developed a fusion-based aircraft detection algorithm that takes advantage of image-based and LiDAR-based detection algorithms \cite{lee2022sensor}. We efficiently combine both algorithms by utilizing conditional random field (CRF) optimization to merge the over-segments occurring in LiDAR detection. In the first step of merging, we project each over-segment onto an image and classify the over-segments projected onto the aircraft bounding box as the primary candidates. We then construct an undirected graph G with the candidates as the vertices and the correlations between vertices as the edges. CRF optimization is used to find the optimal solution, the best label set, by minimizing the graph’s energy function. The graph's energy function that comprises unary potentials and the clique potentials is as follows:
\begin{equation} \label{eq:energy func}
    \Phi(L|S)=\sum_{l_i \in L} U(l_i|s_i)+\sum_{C \in cl(G)}\varphi(l_C|s_C)
\end{equation}
where $L=\{l_1,...,l_k\}$ is the entire class label set, $S=\{s_1,...,s_k\}$ is the object segment, $U(l|s)$ is unary potential defined by the probabilities of the camera and LiDAR detection for each segment, $\varphi(l|s)$ is clique potential defined by the relative relations between adjacent segments, and $C$ represents the clique set of graph $G$. 

\begin{figure}[tp]
    \centering
    \vskip 0.07in
    \includegraphics[width=0.9\columnwidth]{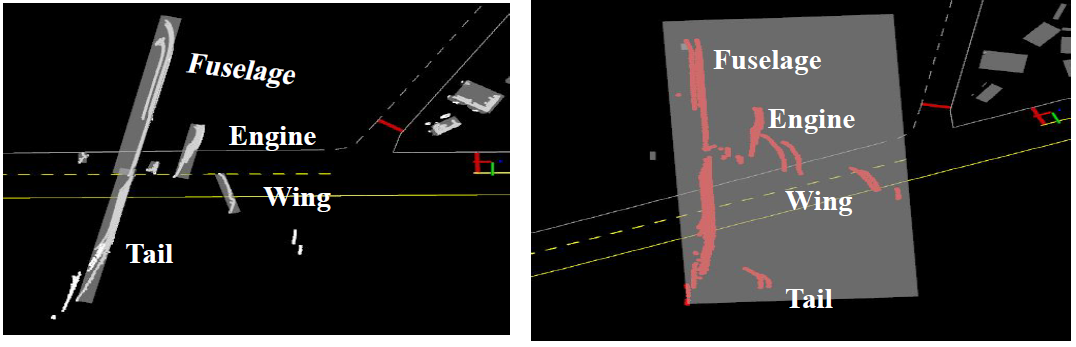}
    \vskip -0.1in
    \caption{Over-segmented aircraft detection results (left) and ours (right).}
    \label{fig:aircraft detection}
    \vskip -0.2in
\end{figure}

Our experiments confirmed that our fusion-based algorithm outperforms other algorithms in terms of aircraft tracking and classification accuracy. We also observed that our algorithm is less affected by changes in aircraft size and has a higher environment-invariant ability compared to other algorithms. These experiments were based on the dataset acquired at the CVG and Dayton International Airport. Fig. \ref{fig:aircraft detection} shows the merged over-segments of aircraft.

\vskip 1mm
\subsubsection{Probabilistic Aircraft Intention Prediction}
Aircraft have top traffic priority in airfields. When aircraft move to the runway from the terminal, a GSE must stop and wait. The GSE also must stop when the wing walker controls the GSE movement before the aircraft moves. The key element in dealing with these situations is the estimation of both aircraft’s pose and its intention. Existing intention prediction algorithms were, however, not appropriate for aircraft because they focused on predicting human or vehicle behavior, which considerably differs from aircraft driving characteristics \cite{hu2018probabilistic}.

To address the above challenge, we developed a probabilistic algorithm that can predict the aircraft’s intention based on the possible situational context information related to the aircraft movement \cite{lee2022probabilistic}. The situational context information ($I$) is comprised of two pieces of evidence: aircraft evidence ($A$) and surrounding object evidence ($O$). The aircraft’s pose and beacon signal are defined as $A$. The information of the GSE and the ramp agent who supports the airport ramp operation are defined as $O$. Each piece of information is expressed as a Bayesian network, which is a probabilistic graphical model. Finally, the aircraft’s intention $\kappa$ is estimated using a Bayesian classifier. The joint distribution of the aircraft intention and the situational context information is presented as follows:
\begin{equation} \label{eq:joint distribution}
    P(\kappa,I|\Theta)=P(\kappa|\Theta)P(A|\kappa,\Theta)P(O|\kappa,\Theta),
\end{equation}
where $\Theta$ is the training parameter set. Likelihood $P(A|\kappa,\Theta)$ and $P(O|\kappa,\Theta)$ are defined based on the statistical distribution of the training dataset. We evaluated our algorithm under various weather conditions and airfield environments, thereby successfully confirming its superior performance compared to other algorithms. Our algorithm demonstrated performance improvements of 2.3 \% in cloudy weather, 1.6 \% in rainy weather, and 9.3 \% in different airfield environments compared to the baseline.

\vskip 1mm
\subsubsection{Efficient Occlusion-aware Risk Assessment via Simplified Reachability Quantification}
In airfields, large objects such as aircraft and K-loaders frequently generate occlusion areas. Occlusion areas are substantially dangerous for autonomous tow tractors because undetected vehicles or workers can suddenly appear on a route. However, conventional risk assessment algorithms do not take into account the occlusion areas. Several recent works have proposed occlusion-aware risk assessment algorithms, but they either fail to identify the occlusion areas in real-time or assess the occlusion risk too conservatively \cite{gilroy2019overcoming}. This can severely hinder the safety and operation efficiency of autonomous tow tractors in airfields.

To overcome the above limitations, we introduced an efficient occlusion-aware risk assessment algorithm that can handle diverse occluded scenarios in airfields \cite{park2023occlusion}. The key component of our algorithm is the simplified reachability quantification that uses a simple distribution model over the occluded object states to estimate their risk. This model-based quantification ensures the constant time complexity of our algorithm. Furthermore, our algorithm can be easily integrated with any planning algorithm by leveraging the additional velocity constraints. We evaluated our occlusion-aware risk assessment under various simulated and real-world occlusion scenarios. The experimental results demonstrated that our algorithm achieves lower collision rates and discomfort scores with lower computation complexity than those of state-of-the-art risk assessment algorithms.

\subsection{Crowded Indoor Environment}
We have been operating an indoor autonomous shuttle service at ICN. The autonomous shuttle called AirRide is required to perform efficient and comfortable driving behaviors in crowded environments. Moreover, AirRide should be implemented with a fail-safe system to ensure passenger safety in various corner cases. The service routes are 300 m at the arrival hall of Terminal 1 and 550 m at the departure hall of Terminal 2. Figs. \ref{fig:environment}b and \ref{fig:vehicle platform}b show AirRide and the served routes, respectively. We conducted a performance evaluation with a total distance of 67.5 km and an operation time of 19.2 h. Table \ref{table:indoor test statistics} presents the results of the performance evaluation.

\begin{table}[hbt!]
    \centering
    \caption{AirRide Autonomous Driving Performance Evaluation.}
    \label{table:indoor test statistics}
    \vskip -0.05in
    \renewcommand{\arraystretch}{1.1}

    \resizebox{\columnwidth}{!}{%
    \small
        \begin{tabular}{c | c | c | c | c}
            \hline 
            \multicolumn{2}{c|}{Perception accuracy} &
            \multirow{2}{*}{
                \begin{tabular}[c]{@{}l@{}}
                    Localization \\ 
                    error
                \end{tabular}
            } &
            \multirow{2}{*}{Driving range} &
            \multirow{2}{*}{
                \begin{tabular}[c]{@{}l@{}}
                    Consecutive \\ 
                    operation time
                \end{tabular}
            } \\ 
            \cline{1-2}
            
            Precision & Recall &  &  &  \\
            \hline
            
            0.86 & 0.88 & 0.06 m & 33.9 km & 9.6 h \\
            \hline
        \end{tabular}%
    }
    \vskip -0.1in

\end{table}

\begin{figure*}[tp]
    \centering
    \includegraphics[width=0.9\textwidth]{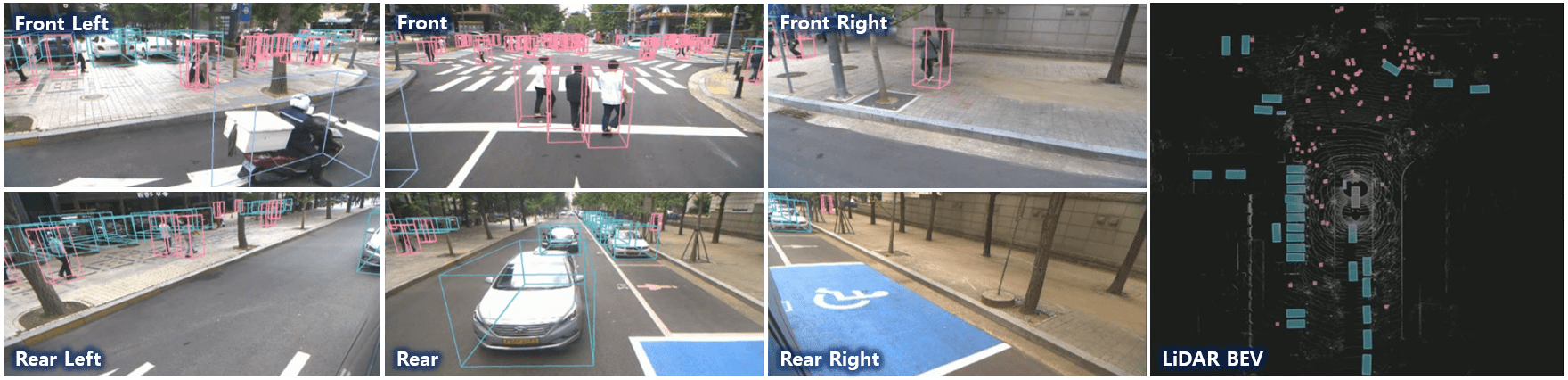}
    \caption{Object detection results in complex urban environments. The cyan, pink, and purple boxes show a vehicle, a pedestrian, and a cyclist, respectively.}
    \label{fig:perception od result}
    \vskip -0.15in
\end{figure*}

\vskip 1mm
\subsubsection{Complex and Narrow Area Collision-free Trajectory Planning}
Autonomous driving in crowded indoor environments requires trajectory optimization algorithms to infer trajectories keeping safe distances from the surrounding dynamic objects. Furthermore, the passages that autonomous vehicles should navigate in indoor environments are much narrower than standard roads. These challenges are fatal to most previous trajectory optimization algorithms that cannot consider both constraints of collision avoidance and kinodynamics in real-time. B-spline-based trajectory optimization is a promising approach that can overcome the limitations of previous algorithms. However, applying previous B-spline-based algorithms to our indoor shuttle is not straightforward, as it has much more complex kinodynamic constraints than other types of vehicles, such as unmanned aerial vehicles.

To address the limitation of previous works, we proposed a new B-spline-based trajectory optimization algorithm that can consider both collision avoidance and kinodynamic constraints in real-time \cite{choi2023safe}. The key idea behind our algorithm is iteratively flattening the collision points of the autonomous vehicle along with pushing penalties from nearby obstacles. To generate these collision points, we also introduce a new swept volume algorithm that can prevent corner-cutting \cite{deits2015efficient} and risk overestimation problems \cite{li2023embodied}, which are common in most previous works. The experimental results showed that our algorithm can generate safe and efficient trajectories on diverse simulated and real-world driving scenarios, thereby achieving a substantial performance gain over baselines, including state-of-the-art trajectory optimization algorithms.

\vskip 1mm
\subsubsection{Fail-safe System for Passenger Safety}
AirRide is not for a short-term proof of concept project but for an actual indoor service. Therefore, it should be able to handle various safety issues that may occur. AirRide is equipped with different fail-safe functions to ensure the safety of passengers and people around the vehicle. When AirRide detects a hardware or software error, it operates a basic fail-safe function to stop and cannot be re-operated until all these errors are resolved. In addition to the basic fail-safe function, we installed service and operation perspective fail-safe functions. First, AirRide can automatically detect sensor occlusion caused by humans covering the sensors or foreign objects. In addition, it uses the seat pressure sensor to determine if passengers are standing. This is primarily due to the risk of passengers occluding the camera or LiDAR (Fig. \ref{fig:vehicle platform}b). Next, AirRide is programmed to avoid entering irrecoverable or irreversible areas where our transportation service cannot be provided, even if some people attempt to force it into these areas intentionally. The fail-safe function for detecting such areas is based on whether AirRide can generate a trajectory to return to the reference path from its current position.

\subsection{Urban Environment}
As shown in Figs. \ref{fig:environment}c and \ref{fig:vehicle platform}c, we have been operating various autonomous driving services, including last-mile delivery, health care, and infra monitoring, in urban environments. Here we describe the novel algorithms we proposed for product-level autonomous shuttle and taxi services in a 31.8 km route in Jinhae, South Korea. The shuttle operates between 19 stops and the taxi operates at 23 stops. These services require real-time and high-accuracy perception performance, and safe and interactive driving behaviors. We conducted a performance evaluation with a total distance of 110.3 km, without any disengagement. Table \ref{table:urban test statistics} presents the autonomous shuttle test statistics in Jinhae.

\subsubsection{Imagination-augmented Hierarchical Reinforcement Learning for Safe and Interactive Motion Planning}
Many previous works demonstrated that rule-based motion planning algorithms can perform safety-aware behaviors following traffic rules in urban environments \cite{broggi2015proud, ziegler2014making}. However, as these algorithms are pre-scripted for specific scenarios and cannot understand interactions with human drivers, their behaviors are often much inferior to the behaviors of expert drivers. While learning-based motion planning is a promising approach for training autonomous vehicles to understand interactions with human drivers, it does not necessarily generate safety-aware driving behaviors.

\begin{table}[tp]
    \centering
    \caption{Autonomous Shuttle Test Statistics in Jinhae.}
    \label{table:urban test statistics}
    \vskip -0.05in
    \renewcommand{\arraystretch}{1.1}

    \resizebox{\columnwidth}{!}{%
        \begin{tabular}{c | c | c | c | c}
            \hline 
            Stations & Roundabouts &
            \begin{tabular}[c]{@{}l@{}}
                Unsignalized \\ 
                intersections
            \end{tabular} &
            \begin{tabular}[c]{@{}l@{}}
                Lane \\ 
                changes
            \end{tabular} &
            Barricades
            \\
            \hline 
            141 & 25 & 525 & 85 & 10 \\
            \hline
        \end{tabular}%
    }
    \vskip -0.2in

\end{table}

To achieve safe and interactive driving behaviors in urban environments, we proposed a new motion planning algorithm called imagination-augmented hierarchical reinforcement learning (IAHRL) \cite{lee2023imagination}. IAHRL efficiently combines the strengths of rule-based and learning-based motion planning algorithms. The key idea behind IAHRL is that the low-level policies imagine safe and structured behaviors, and then the high-level policy is trained to infer interactions with surrounding vehicles by interpreting the imagined behaviors. The low-level policies are implemented with a rule-based motion planning algorithm introduced by Werling \textit{et al.} \cite{werling2010optimal}. We also introduce a new attention mechanism that allows our high-level policy to be permutation-invariant to the order of surrounding vehicles and prioritize an ego vehicle over them. It enables the ego vehicle to perform stable and robust behaviors in high-traffic scenarios. The attention mechanism can be defined as follows: 
\begin{equation} \label{eq:attention structure}
    m_t = \sigma(Q(\zeta_t^{ego},\eta) K(s_t)^T)\:V(s_t),
\end{equation}
where $\sigma$ is a softmax function, $Q$, $K$ and $V$ are the query, key, and value matrices, respectively, $s_t$ is a state, $\zeta_t^{ego}$ is the imagined behavior of the ego vehicle, and $\eta$ is a random seed vector. The state consists of imagined behaviors of the ego vehicle and surrounding vehicles $s_t=\{\zeta_t^{ego}, \zeta_t^1, \zeta_t^2, \dots, \zeta_t^n\}$. Note that the key and value matrices are linear transformations of $s_t$, and the query matrix is linear transformations of $\zeta_t^{ego}$. The experimental results showed that IAHRL achieves success rates of over 95 \% in various unsignalized intersections and lane-change tasks, significantly outperforming baselines. We also observed that IAHRL enables an autonomous vehicle to give its attention to surrounding vehicles in a manner akin to human drivers.

\begin{figure*}[tp]
    \centering
    \includegraphics[width=0.85\textwidth]{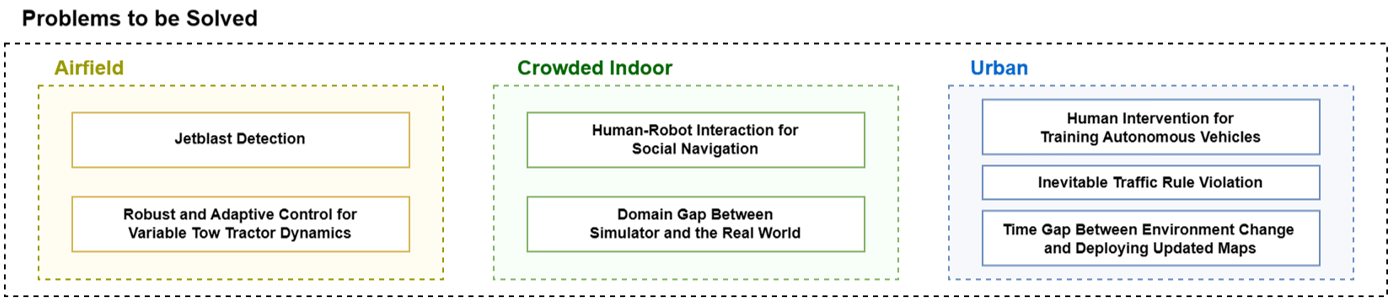}
    \vskip -0.1in
    \caption{Problems to be solved for each environment.}
    \label{fig:problems to be solved}
    \vskip -0.1in
\end{figure*}

\vskip 1mm
\subsubsection{Real-time and Accurate Multi-sensor Object Detection}
Object detection algorithms for autonomous vehicles must ensure real-time and high accuracy with limited computation capacity. Cross modal transformer (CMT) \cite{yan2023cross}, a state-of-the-art sensor fusion algorithm, showed a high performance by utilizing transformers to fuse the features of LiDAR and multiple cameras. Nevertheless, it requires high computational resources, making it difficult to satisfy the real-time constraints of autonomous vehicles.

We developed a multi-sensor object detection (MSOD) algorithm that achieves high performance while satisfying real-time constraints. MSOD extracts bird’s-eye view features from LiDAR data and employs only the corresponding image features as the transformer input. Thus, it can more efficiently operate than the CMT, which uses all LiDAR and camera features as the input. MSOD uses the model to leverage the advantages of each sensor by utilizing LiDAR features for depth information and camera features for semantic information, ensuring high accuracy. Another distinctive feature of MSOD is its ability to operate with LiDAR alone even under camera failure.

Open datasets, such as KITTI  and nuScene, differ from the Korean urban environment; hence, we generated a dataset using our autonomous driving platform (Fig. \ref{fig:vehicle platform}) to accurately evaluate the performance. The evaluation metrics were nuScene detection score (NDS) and mean average precision (mAP). We compared inference speed tested on RTX3090 GPU using Python. Fig. \ref{fig:perception od result} and Tabel \ref{table:perception od evaluation} present the qualitative and quantitative results, respectively. The experimental results indicate that our algorithm, while maintaining accuracy comparable to CMT, is nearly twice as fast. We confirmed that MSOD can operate at a frequency of 28 Hz after conversion to TensorRT.

\begin{table}[hbt!]

    \centering
    \caption{Object detection performance evaluation.}
    \label{table:perception od evaluation}

    \vskip -0.05in

    \resizebox{0.95\columnwidth}{!}{%
    \scriptsize
        \begin{tabular}{c || c | c c | c}
            \hline
            Methods & Modality & NDS $\uparrow$ & mAP $\uparrow$ & FPS $\uparrow$ \\
            \hline\hline
            CMT \cite{yan2023cross} & LiDAR, camera & 0.6758 & 0.582 & 3 \\
            \hline
            MSOD-L & LiDAR & 0.6533 & 0.532 & 8 \\
            MSOD & LiDAR, camera & 0.6736 & 0.630 & 5.7 \\
            \hline
        \end{tabular}%
    }

    \vskip -0.15in

\end{table}

\section{Problems to be Solved}
This section introduces the problems that have not received much attention but must be solved to advance the autonomous driving services for each environment (Fig. \ref{fig:problems to be solved}).

\subsection{Airfield Environment}
\subsubsection{Jetblast Detection}
Jetblast is an intangible rapid airflow from an aircraft’s engines and typically occurs when it prepares for takeoff or right after landing. Autonomous tow tractors should be able to avoid the jetblast, as it can cause significant damage to them. Unfortunately, most previous detection algorithms for autonomous driving have focused on detecting objects common in urban environments, such as vehicles and pedestrians. We observed that these algorithms are inefficient in detecting the jetblast due to its unique features. To overcome this challenge, we are currently developing a thermal camera-based jetblast detection algorithm (Fig. \ref{fig:airfield problems}). Thermal cameras are widely used to detect such intangible objects. We confirmed that jetblast is successfully detected using a YOLO-based 2D detection network with temperature features. Moreover, we believe that incorporating sound localization through audio detection can further enhance the jetblast detection accuracy \cite{tran2021audio}.

\vskip 1mm
\subsubsection{Robust and Adaptive Control for Variable Tow Tractor Dynamics}
\begin{figure}[tp]
    \centering
    \vskip 0.07in
    \includegraphics[width=0.77\columnwidth]{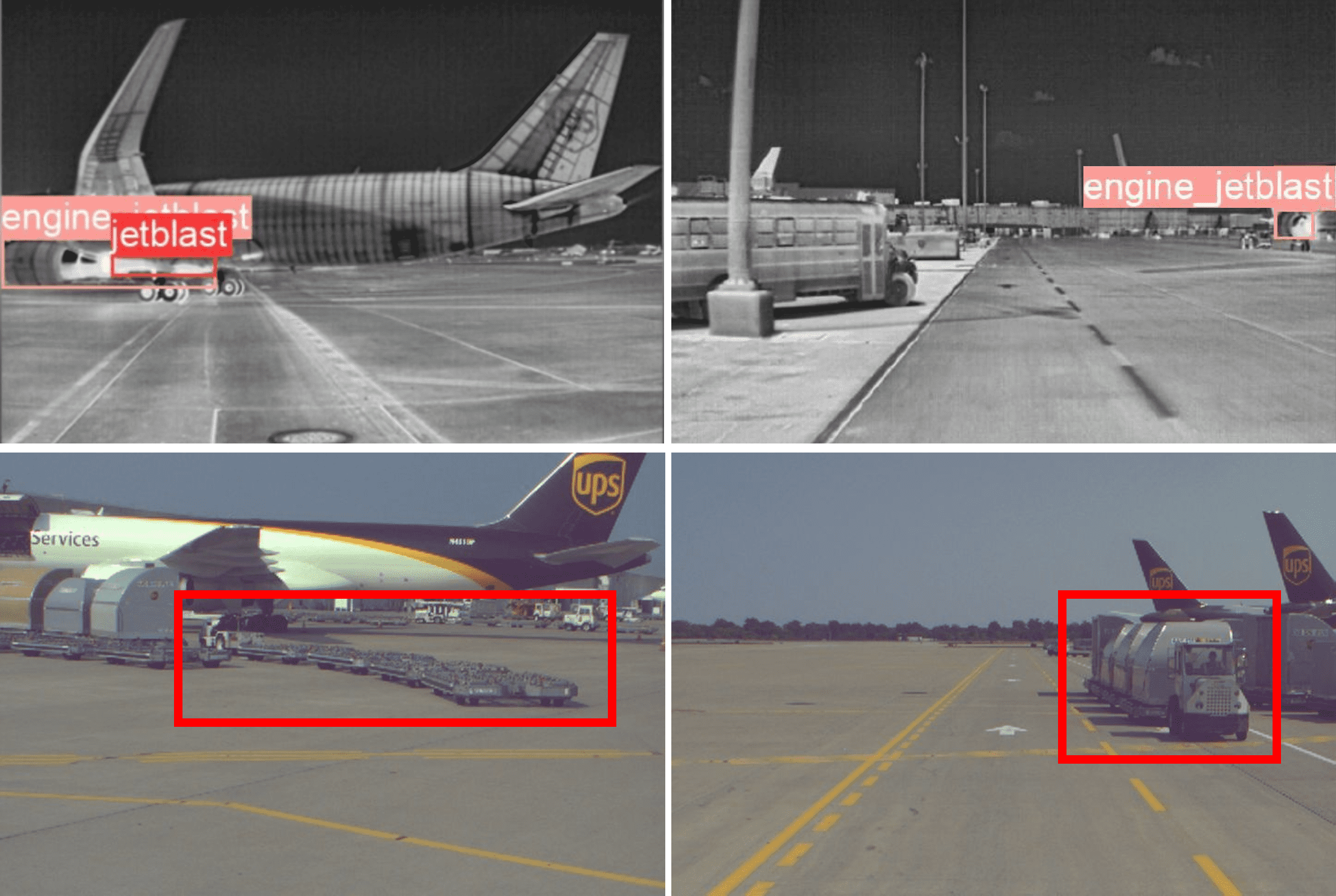}
    \caption{Top: jetblast detection with a thermal camera. Bottom: tow tractor’s payload differences: empty (left) and fully loaded (right).}
    \label{fig:airfield problems}
    \vskip -0.15in
\end{figure}

The tow tractor's payload can change as shown in Fig. \ref{fig:airfield problems}, and it significantly affects the dynamics of tow tractors. To ensure the efficiency and safety of autonomous tow tractors, a robust control algorithm that can adapt to changing dynamics is essential. Most previous control algorithms for tow tractors handle this challenge by assuming a prior knowledge of the tractor's configuration. However, the tractor's configuration varies frequently and cannot be predicted in advance. Meta-learning that facilitates effective and continuous online adaptation is a promising approach for handling the changing dynamics problem \cite{finn2017model}. Nagabandi \textit{et al.} \cite{nagabandi2018learning} introduced a model-based meta-learning algorithm that can continuously adapt a model to recent transitions in dynamic environments. We believe that integrating the benefits of meta-learning into our controller can enable tow tractors to adapt to changing dynamics due to payload.

\subsection{Crowded Indoor Environment}
\subsubsection{Human-Robot Interaction for Social Navigation}
In general, navigation algorithms for indoor robots should infer efficient and adaptive paths that do not interrupt people's behaviors \cite{medina2022perception}. AirRide can generate such paths by predicting people's behaviors in advance.
However, AirRide was occasionally frozen when people standing in line for boarding blocked whole passages, which is common at arrival and departure halls. While AirRide is programmed to stop and then request yielding through voice guidance when it encounters these scenarios, we observed that our voice guidance was not sufficient to enable AirRide to encourage people to cooperate. We believe that investigating socially compliant driving is a good starting point for handling our limitation \cite{chen2019crowd}. We thus will analyze how existing socially compliant driving algorithms work in the challenging scenario at the arrival and departure halls. 

\vskip 1mm
\subsubsection{Domain Gap Between Simulator and the Real World}
A simulator is essential for evaluating autonomous vehicles before deploying them in real-world environments. While numerous simulators have been developed for urban environments and have achieved high fidelity \cite{dosovitskiy2017carla}, those designed for crowded indoor environments still have many shortcomings \cite{kaur2022simulators}. One of the main limitations is the inaccurate modeling of people's diverse behaviors, such as pulling their luggage and pushing carts at airports. Specifically, behaviors resulting from people's interest in autonomous vehicles have been overlooked in most simulators. Camara \textit{et al.} \cite{camara2020pedestrian} introduced realistic modeling of people's behavior, ranging from simple individual prediction to game-theoretic models of interactions between people and autonomous vehicles. This realistic modeling can significantly reduce the domain gap between simulators and real-world environments, which is crucial for advancing autonomous driving research in crowded indoor environments.

\subsection{Urban Environment}

\subsubsection{Human Intervention for Training Autonomous Vehicles}
While IAHRL introduced in Section \uppercase\expandafter{\romannumeral3}.\textit{C.1)} enables autonomous vehicles to learn safe and interactive driving behaviors in urban environments, a lot of additional challenges should be addressed to truly facilitate the training of autonomous vehicles in real-world environments. One challenge that has not received much attention is the substantial human intervention and supervision required for training autonomous vehicles. Humans should continuously supervise autonomous vehicles during training to intervene before they enter unsafe states. Furthermore, after each episode, humans should reset an environment by manually driving autonomous vehicles to the initial states for the next episode. Autonomous RL (ARL), which learns not only how to solve a task but also how to reset an environment, is a promising approach to minimizing human intervention and supervision. Several recent works introduced impressive ARL algorithms that can substantially reduce human intervention in diverse tasks \cite{sharma2021autonomous}. However, applying these algorithms to train an autonomous vehicle in real-world environments is infeasible, as they cannot guarantee safety-aware reset behaviors required in autonomous driving tasks. We believe that developing ARL algorithms that can safely reset an environment is crucial for realizing autonomous vehicles in the real world.

\vskip 1mm
\subsubsection{Inevitable Traffic Rule Violation}
Autonomous vehicles should follow traffic rules to coexist with human-driven ones. However, we observed that autonomous vehicles should occasionally violate a subset of traffic rules to handle several urban driving scenarios. For instance, faced with illegally parked vehicles blocking the ego lane, autonomous vehicles may need to cross the centerline to overtake. Autonomous vehicles may also cross the stop line and stop if the traffic signals unexpectedly change. Autonomous vehicles cannot handle these scenarios because they are programmed to follow traffic rules. A promising approach to overcoming this limitation is to mimic the driving behaviors of expert drivers who can handle these scenarios by following mutually accepted rules while minimally violating the traffic rules. Cho \textit{et al.} \cite{cho2018learning} encoded traffic rules with a signal temporal logic and utilized Gaussian process regression to learn when to violate traffic rules from expert demonstrations. Lee \textit{et al.} \cite{lee2017learning} used inverse reinforcement learning to infer the cost functions representing the driving behaviors of expert drivers. We believe that further research on this approach will be an important step toward human-like autonomous vehicles.

\vskip 1mm
\subsubsection{Time Gap Between Environment Change and Deploying Updated Maps}
High-definition (HD) maps are used in various components of autonomous driving systems, including map matching-based localization, trajectory prediction, and global path planning algorithms. Outdated HD maps can significantly degrade autonomous driving performance. However, the continuous and autonomous updating of HD maps to reflect environmental changes is not straightforward in the real world. To overcome this challenge, Kim \textit{et al.} \cite{kim2021hd} introduced a crowd-sourcing algorithm that leverages a large number of vehicle fleets with low-cost sensors to continuously update the maps in real-time. However, even with the fast map update through crowd-sourcing, there is a time gap between environment change and deploying updated maps to autonomous vehicles. A fail-operational system, maintaining continuous operation during system failures, is one of the promising solutions that can handle this time gap, but it has yet to receive much attention. We believe that switching to a perception-based autonomous driving mode, which relies solely on current perception results without HD maps, when environment change is detected, is an appropriate fail-operational system.

\section{Conclusion}
We share our experimental results and lessons obtained from autonomous driving services in airfield, crowded indoor, and urban environments. Each environment has its own unique challenging problems and we introduce general and novel algorithms for addressing them. We also provide an overview of promising research directions for the remaining problems that have not received much attention but must be addressed. Exploring these research directions is crucial for realizing the massive potential of autonomous vehicles. We hope that our work will be helpful to researchers who would like to deploy autonomous vehicles in various real-world environments.

\bibliographystyle{IEEEtran}
\bibliography{IEEEabrv,reference}

\end{document}